\documentclass{article}
\usepackage{arxiv}
\usepackage[utf8]{inputenc} 
\usepackage[T1]{fontenc}    
\usepackage{hyperref}       
\usepackage{url}            
\usepackage{booktabs}       
\usepackage{amsfonts}       
\usepackage{nicefrac}       
\usepackage{microtype}      
\usepackage{lipsum}		
\usepackage{graphicx}
\usepackage{natbib}
\usepackage{doi}
\usepackage{algorithm}
\usepackage{algorithmic}

\title{WEPO: Web Element Preference Optimization for LLM-based Web Navigation}

\author{ Jiarun Liu, Jia Hao, Chunhong Zhang, Zheng Hu \\
    State Key Laboratory of Networking and Switching Technology\\
	Beijing University of Posts and Telecommunications, Beijing 100876, China \\
	\texttt{\{liujiarun01, zhangch, huzheng\}@bupt.edu.cn} \\
}

\date{}

\renewcommand{\headeright}{}
\renewcommand{\undertitle}{}
\renewcommand{\shorttitle}{WEPO: Web Element Preference Optimization for LLM-based Web Navigation}

\hypersetup{
pdftitle={WEPO: Web Element Preference Optimization for LLM-based Web Navigation},
pdfsubject={},
pdfauthor={Jiarun Liu},
pdfkeywords={Web Navigation, Preference Learning},
}

\begin{document}
\maketitle

\begin{abstract}
The rapid advancement of autonomous web navigation has significantly benefited from grounding pretrained Large Language Models (LLMs) as agents. However, current research has yet to fully leverage the redundancy of HTML elements for contrastive training. This paper introduces a novel approach to LLM-based web navigation tasks, called Web Element Preference Optimization (WEPO). WEPO utilizes unsupervised preference learning by sampling distance-based non-salient web elements as negative samples, optimizing maximum likelihood objective within Direct Preference Optimization (DPO). We evaluate WEPO on the Mind2Web benchmark and empirically demonstrate that WEPO aligns user high-level intent with output actions more effectively. The results show that our method achieved the state-of-the-art, with an improvement of 13.8\% over WebAgent and 5.3\% over the visual language model CogAgent baseline. Our findings underscore the potential of preference optimization to enhance web navigation and other web page based tasks, suggesting a promising direction for future research.
\end{abstract}


\section{Introduction}

\begin{figure*}[t]
    \includegraphics[width=\linewidth]{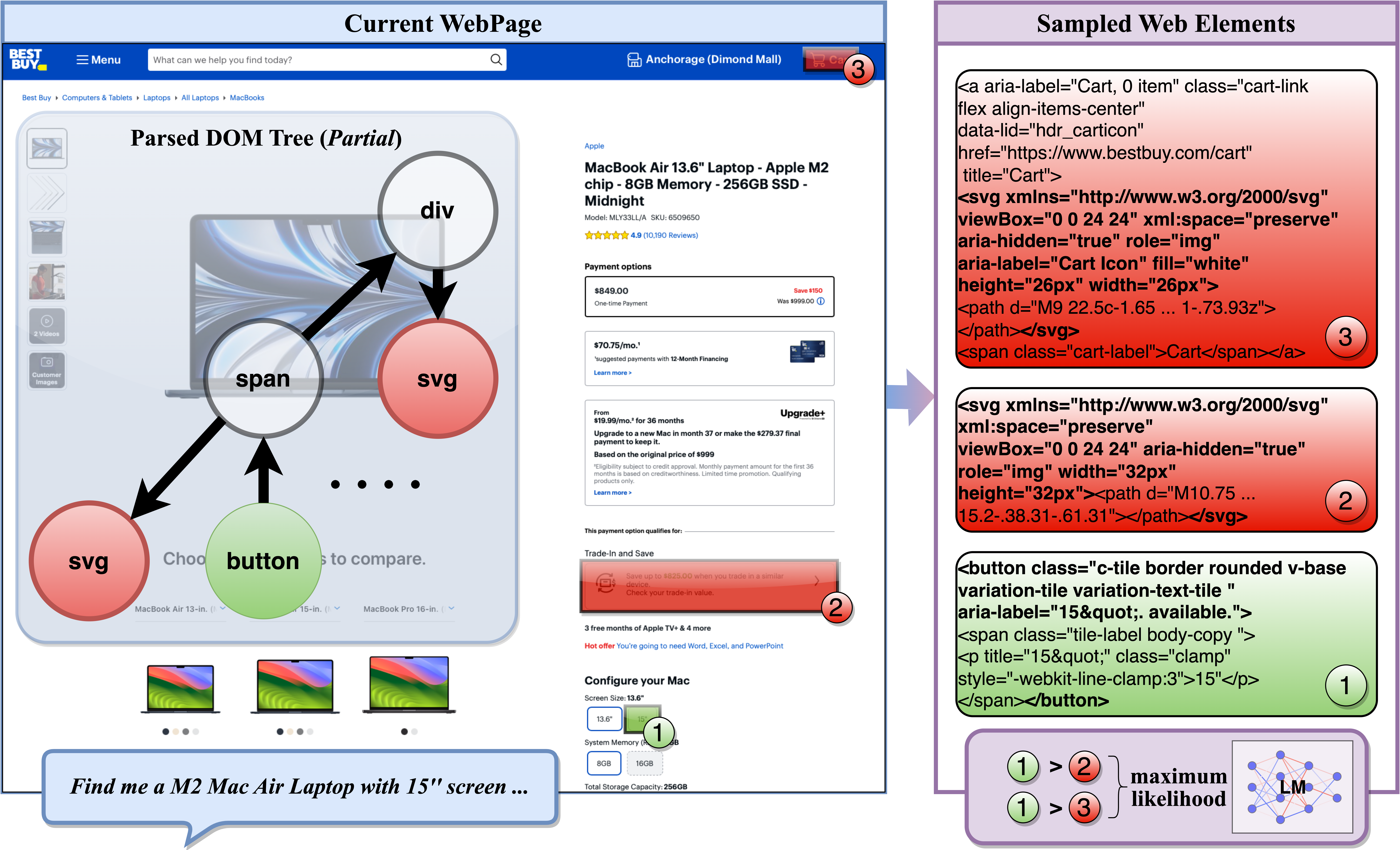}
    \caption{Illustration of Web Element Preference Optimization (WEPO). Given user intent, \textit{Find me a M2 Mac Air Laptop with 15" screen}, WEPO combines the correct element (marked in green) with heuristic rule-based sampled negative elements (marked in red) to construct preference pairs. This process utilizes the maximum likelihood objective function proposed in algorithms such as DPO to fine-tune the language model, thereby enhancing its accuracy in element discrimination and selection.
  }
    \label{fig:intro}
  \end{figure*}

  The field of autonomous web navigation has seen significant advancements, driven by the capabilities of Large Language Models (LLMs) in both mobile and webpage interactions \citep{wang2024survey,mialon2023augmented,xi2023rise}. Preliminary attempts, such as the ChatGPT Plugin \citep{openai2023chatgptplugins}, have also started building practical applications of web knowledge-based chatbot. 

  Web navigation can be described as processes where agents perform specific tasks on behalf of human users within a web environment, involving the interpretation of high-level user instructions, decomposing them into basic operations, and interacting with complex web pages dynamically. To achieve this, agents must understand intricate web scenarios, adapt to dynamic changes such as noisy text and evolving HTML structures, and generalize successful operations to unseen tasks, thus freeing humans from repetitive interactions with computer interfaces.
  
  Traditional web agents trained through reinforcement learning \citep{shi2017world,yao2022webshop} often mimic human behavior using predefined actions like typing, searching, and navigating to a specific page. However, they frequently struggle with the complexities of real-world web environments and the challenges of designing effective reward functions. Recent research has leveraged the HTML understanding, logical reasoning, and code generation capabilities of LLMs, enabling agents to comprehend long HTML documents and predict the next action steps. Notable examples include Mind2Web \citep{deng2024mind2web}, which provides an realistic interaction dataset and fine-tunes multiple LLMs to summarize verbose HTML and iteratively optimize and execute actions. Other works such as WebGum \citep{furuta2023multimodal} and CogAgent \citep{hong2023cogagent} construct multimodal architectures, enhancing agents with visual perception abilities through supervised learning with a multimodal corpus that includes HTML screenshots. These prior works are thoroughly summarized in our related work section.
  
  In parallel, preference learning in fine-tuning LLMs has gained prominence, particularly since the introduction of Reinforcement Learning with Human Feedback (RLHF) in GPT-3 \citep{ziegler2019fine,ouyang2022training}, which aligns model outputs with human preferences through reward modeling and reinforcement learning with KL divergence constraints. More subsequent works, such as Direct Preference Optimization (DPO) \citep{amini2024direct} and its variants \citep{hong2024orpo,morimura2024filtered}, have reparameterized the reward function and optimized training efficiency. These advancements have primarily focused on mainstream tasks in natural language processing, such as dialogue generation and code generation. To our knowledge, the proven effectiveness of preference optimization algorithms like DPO has not been applied to web task automation.
  
  Moreover, existing autonomous web navigation research has not fully exploited the potential of contrastive learning using non-salient HTML elements. After observing the structural complexity and crowded element arrangement in HTML on both Mind2Web and real web environment, we realized that these environments are naturally conducive to data-augmented preference learning. We hypothesize that incorporating preference learning can significantly enhance LLM-based agents' capabilities in web navigation tasks. 
  
  Motivated by this, we introduce Web Element Preference Optimization (WEPO), a novel framework that integrates preference optimization algorithms into mainstream LLM-based web navigation tasks. By sampling non-relevant web elements as negative samples, we implement preference learning that requires no human effort, thereby utilizing redundant information in the web environment and achieving high sample efficiency. Specifically, we design a heuristic distance-based element sampling method tailored to the DOM tree structure to enhance the efficiency of contrastive learning. WEPO then maximizes the likelihood of operations on preferred elements and minimizes it for dis-preferred elements, aligning user high-level intent with agent operation sequences. We illustrate WEPO in Figure \ref{fig:intro}, provide detailed implementation steps and the theoretical foundation of WEPO in the subsequent sections.
  
  For experiments, we selected the Mind2Web dataset due to its high task diversity and realistic web scenarios, which best validate the capabilities of fine-tuned LLM agents. Our experiments on multiple mainstream open-sourced models demonstrate that our WEPO significantly outperforms traditional supervised fine-tuning (SFT) methods, exceeding the MindAct \citep{deng2024mind2web} baseline by $20.0\%$ and WebAgent \citep{gur2023real} by $13.8\%$. WEPO also surpasses visual language model (VLM) CogAgent \citep{hong2023cogagent} by $5.3\%$ with smaller model parameters and faster inference time, achieving state-of-the-art performance.
  
  We believe WEPO represents a significant advancement in autonomous web navigation, leveraging HTML structure based preference optimization to enhance task performance and suggesting promising directions for future research in LLM-based web navigation and related applications.

\section{Related Work}

  \textbf{Web Navigation with LLMs.} In the area of web navigation and web-based task automation, the integration of LLM-based agents has shown considerable promise. \citet{kim2024language} and \citet{sridhar2023hierarchical} introduces the use of prompting schemes combined with criticism or hierarchical modularized design, \citet{li2023zero} exploits zero-shot prompt learning via self-reflection and structured thought management, and \citet{zheng2023synapse} applies structural prompting with exemplar retrieval to achieve few-shot in-context learning.
  
  \citet{gur2023real} and \citet{gur2023understanding} demonstrated the effectiveness of encoder-decoder architectures such as HTML-T5, tailored to the HTML tree structure through sophisticated local and global attention mechanisms and a mixture of denoising objectives. \citet{deng2024mind2web} introduced MindAct framework, which simplifies web interaction by transforming generation tasks into multiple-choice formats through instruction fine-tuning, thereby gaining decent performance on Mind2Web benchmark. In addition, \citet{nakano2021webgpt} developed WebGPT, which leverages reinforcement learning with human feedback (RLHF) \citep{ouyang2022training,ziegler2019fine} to align decision-making processes with human-preferred answers. \citet{yao2022webshop} discusses the integration of reinforcement learning (RL) for scalable real-world web shopping, highlighting the design of heuristic rewards that enhance learning efficiency. CC-Net \citep{humphreys2022data}, despite not using large language models, utilizes a hybrid architecture that integrates pixel-based inputs via ResNet \citep{he2016deep} blocks and language embeddings through transformer blocks \citep{vaswani2017attention}, demonstrating exceptional performance through its effective combination of RL and imitation learning.Attempts to address web navigation using multimodal large language models (MLLMs) are also emerging at scale \citep{hong2023cogagent,you2024ferret,wang2024mobile,niu2024screenagent,baechler2024screenai,cheng2024seeclick,furuta2023multimodal}. Among them, CogAgent once reached the state-of-the-art on Mind2Web benchmark by using a high-resolution cross-modular image encoder in conjunction with visual language model (VLM).
  
  Benchmarks in web navigation have evolved rapidly from the simplified MiniWoB \citep{shi2017world} to the advanced Mind2Web \citep{deng2024mind2web} and other alternatives \citep{lu2024weblinx,zhou2023webarena,he2024webvoyager}. Mind2Web tackles real-world complexities by incorporating 137 real-world websites into a wide range of 2350 multi-step tasks across 31 domains.

  \textbf{Preference Learning with LLMs.} Fine-tuning large language models with preference objective has evolved significantly with Direct Preference Optimization (DPO) \citep{rafailov2024direct} and its variants \citep{hong2024orpo,meng2024simpo,morimura2024filtered,amini2024direct} improving on traditional RLHF approaches \citep{christiano2017deep,ziegler2019fine,stiennon2020learning,ouyang2022training}. Recent advances focus on improving dataset quality \citep{morimura2024filtered}, introducing marginal distinctions to better control bias \citep{duan2024negating}, optimizing the preference labeling process and enhancing sample efficiency by minimizing human oracle involvement \citep{bai2022constitutional} and acquire comparison pairs actively \citep{muldrew2024active}.
  
  \textbf{Preference Learning in Web Scenarios.} The application of preference learning to web-based tasks is not a new concept. Notable early work by \citet{radlinski2005query} leveraged query chains and implicit user feedback, such as click-through data, to refine search engine algorithms. This approach aimed to capture subtle user preferences that were not explicitly stated but could be inferred from their search behavior sequences. \citet{xiang2010context} and \citet{zhu2021contrastive} also explored preference modeling and ranking for web retrieval applications, including data augmentation of user interaction sequences for comparative learning. These works provide a solid foundation for preference in web scenarios, although they predate the era of large language models and did not attempt to address web navigation issues directly. As mentioned above, \citet{nakano2021webgpt} revolves around fine-tuning GPT-3 \citep{brown2020language} to answer long-form questions in a text-based web environment. The training of WebGPT involves behavior cloning followed by rejection sampling against a reward model trained to predict human preferences, which is considered as an adaptation of preference learning to web QA tasks.

  \section{Web Element Preference Optimization}

  \subsection{Task Formulation}
  
  We formulate the web navigation task as a partially observable Markov decision process (POMDP) $(S, A, T, R, I, O)$ according to \citet{yao2022webshop}, with state space $S$, action space $A$, deterministic transition function $T:S \times A \rightarrow S$, reward function $R:S \times A \rightarrow [0,1]$, intent space $I$ and a state observation space $S \times I \rightarrow O$. A state $s \in S$ represents a webpage, an action $a \in A(s)$ corresponds to an operation on a webpage, a high-level natural language intent $i \in I$ represents a complex web interaction, usually involving implicit multi-step sub-instructions. Consistent with mainstream efforts \citep{deng2024mind2web,gur2023real}, we discard the use of a reward function \( r = R(s, a) \) and do not consider employing reinforcement learning in this work. Within interaction loop of the web environment, numerous web elements \( e_k \) exist, yielding candidate set \( E = \{ e_1, e_2, ... , e_m \} \). The target element is denoted as \( \hat{e} \) with a ground truth operation \( \hat{o} \), which are both labeled through human supervision, thereby determining \( \hat{a} = a(\hat{e},\hat{o}) \).
  In the Mind2Web \citep{deng2024mind2web} setting, a state \( s \) includes snapshots in multiple formats, such as HTML code and trace files. We exclusively utilize the HTML scripts of the webpage for WEPO learning. For each state, \( E \) comprises all web elements that collectively form the current page. An action \( a \) encompasses clicking on an interactive element \texttt{(CLICK,element\_ID)}, inputting textual content to an input field \texttt{(TYPE,element\_ID,value)} and selecting an option \texttt{(SELECT,element\_ID,value)}. See more statistical analysis of Mind2Web in \hyperref[sec:stat]{Appendix E}.
  
  \subsection{WEPO Implementation}
  
  We start by introducing the sampling mechanism of WEPO, as illustrated by the partial DOM tree shown in Figure \ref{fig:intro}. Since every webpage can be parsed into a corresponding DOM tree with each web element represented by a unique node, web elements corresponding to nodes that are closer in proximity under the same ancestor within the DOM typically exhibit greater functional and semantic similarity. Building on this characteristic, we have developed a distance-based sampling method specifically tailored to the DOM tree structure, which begins by selecting a substantial number (top \( k \)) of element anchors on the page, including one correct element, and then calculates and sorts the sum of the distances between negative and positive samples to their lowest common ancestor (LCA) to quantify their relative distances. Subsequently, the method proportionally samples the top \( n \) closest elements from the sorted results, which are then combined with the correct element to form comparisons. We aim to enable the model to effectively learn to distinguish between web elements with similar functions based on the given operational intent.

      \begin{figure*}[ht]
      \begin{equation}
        \label{eq:1}
        \mathcal{L}_{\mathrm{DPO}}\left(\pi_{\theta} ; \pi_{\mathrm{ref}}\right)=-\mathbb{E}_{\left(x, a_{w}, a_{l}\right) \sim \mathcal{D}}\left[\log \sigma\left(\beta \log \frac{\pi_{\theta}\left(a_{w} \mid x\right)}{\pi_{\mathrm{ref}}\left(a_{w} \mid x\right)}-\beta \log \frac{\pi_{\theta}\left(a_{l} \mid x\right)}{\pi_{\mathrm{ref}}\left(a_{l} \mid x\right)}\right)\right]
      \end{equation}
      \end{figure*}
  
  During training, we implement Direct Preference Optimization in WEPO, since DPO is free of reward modeling and training stable \citep{rafailov2024direct}. By optimizing the target loss function, WEPO aims to increase the likelihood of operations on preferred elements and decrease the likelihood of operations on dis-preferred elements. As shown in Equation \ref{eq:1}, we introduce the maximum likelihood objective proposed in DPO and adaptively modify the preferred completion \( y_w \) and dis-preferred completions \( y_l \) into preference action pairs \( a_w \) and \( a_l \). Given the pretrained model \( \pi_{\theta} \) and the reference model \( \pi_{\mathrm{ref}} \) initialized from \( \pi_{\theta} \), we fine-tune \( \pi_{\theta} \) according to Equation \ref{eq:1}, where \( \beta \) is a hyperparameter that controls the penalty for deviations from \( \pi_{\mathrm{ref}} \).
  
  We demonstrate the pseudo-code for WEPO implementation in Algorithm \ref{alg:Framework}, which illustrates how \( a_w \), \( a_l \), and \( x \) used for optimizing are obtained at each training step. First, we clean and prune the HTML code. Consistent with previous work \citep{gur2023understanding, deng2024mind2web}, we adapt an element-centric approach to isolate HTML snippets. By focusing on a key element, we navigate its ancestors within the HTML tree, guided by a simple constraint that monitors the tree's width and depth. We stop this traversal once the number of descendants exceeds predefined thresholds, thus defining the snippet using the sub-tree. We introduce a pruning ratio \( k \) that represents the number of target elements remaining after pruning. We ensure that the pruned HTML snippet contains the ground truth element during training; In inference time, we use a small ranking LM derived from the candidate generation stage of \citet{deng2024mind2web} to implement priority-based HTML pruning, which scores all elements first and then selects the top-$k$ elements with the largest logits. Subsequently, we concatenate the preprocessed HTML state \( s' \), historical trajectory \( \tau \), initial intent \( i \) and a sophisticated prompt template \( P \) to generate the input \( x \), where \( \oplus \) denotes string concatenation. See more details of prompt example flow in \hyperref[sec:flow]{Appendix F}.

  \begin{algorithm}
      \caption{WEPO Algorithm} 
      \label{alg:Framework} 
      \begin{algorithmic}[1]
      \REQUIRE (for each step) ~~\\
      Intent $i$, current HTML $s$, action trajectory $\tau$; prompt template $P$;\\
      pretrained LM $\pi_{\theta}$, reference LM $\pi_{\mathrm{ref}}$ and deviation parameter $\beta$;\\
      Pruning ratio $k$, negative ratio $n$;\\
      Target element $\hat{e}$, corresponding operation $\hat{o}$ and target action $\hat{a} = a(\hat{e}, \hat{o})$;\\
      \ENSURE (for each step) ~~\\
      \STATE Clean and prune the HTML DOM tree with $k$ elements remaining ${s}' = f_{prune}(s, k)$;
      \label{ code:fram:prune }
      \STATE Concatenate $x = {s}' \oplus \tau \oplus i \oplus P$
      \STATE Get positive action $a_w = \hat{a}$;
      \STATE Sample $n$ negative elements $\left \{ e_{l_1}, e_{l_2}, ..., e_{l_n} \right \} $ from candidate set $E \leftarrow {s}'$ based on LCA distance from $\hat{e}$;
      \label{ code:fram:sample }
      \FOR{ $i = 1$ to $n$}
      \STATE Set $\epsilon \sim random\_uniform(0,1)$
      \STATE $o_{l_i} \sim f_{op}(\{\texttt{CLICK}, \texttt{TYPE}, \texttt{SELECT}\}, \hat{o}, \epsilon)$
      \STATE Get $i$-th negative action $a_{l_i} = a(e_{l_i}, o_{l_i})$
      \STATE Optimize $\nabla_{\theta} \mathcal{L}_{\mathrm{DPO}}\left(\pi_{\theta}; \pi_{\mathrm{ref}}\right)$ in Equation \ref{eq:1} given $\beta$, $x$, $a_w$ and $a_l = a_{l_i}$;
      \ENDFOR
      \end{algorithmic}
  \end{algorithm}

  Subsequently, we apply the tailored distance-based sampling method to yield candidate set \( E \). We set the number of negative samples as \( n \), corresponding to a positive-to-negative sample ratio of $1:n$. After obtaining the negative elements $\left \{ e_{l_1}, e_{l_2}, ..., e_{l_n} \right \}$, we employ a designed heuristic rule $f_{op}$ to randomly sample the corresponding negative operations $o_{l_i}$, which involves selectively replacing \texttt{TYPE} and \texttt{SELECT} to ensure balanced and diverse sample types. This replacement occurs only when $\hat{o} \ne \texttt{CLICK}$, as the negative samples obtained through sampling have a very low probability of being \texttt{TYPE} or \texttt{SELECT} (see \hyperref[sec:stat]{Appendix E}), adding data balance without confusing the LLM about the functionality of webpage elements. The rule empirically sets the replacement probability threshold at $0.33$, and several verification confirmed that values around this threshold have no significant impact on WEPO. Therefore, when the positive sample is a \texttt{CLICK} operation, the negative samples are also \texttt{CLICK}; however, when the positive sample involves the other two actions, the negative samples might be changed to \texttt{CLICK}. Finally, we obtained \( a_w \), \( a_l \), and \( x \), and used Equation \ref{eq:1} to perform gradient back-propagation on the pretrained LM $\pi_{\theta}$, with $\beta$ controlling the deviation from the reference model $\pi_{\mathrm{ref}}$. We opt for straightforward random sampling as an alternative, where the randomly distributed negative samples also maintain a low correlation with $\hat{e}$ on the webpage. For further discussion on the independence of non-salient sampled elements, please see \hyperref[sec:hypo]{Appendix B}.

  \section{Experiments}
    
  \subsection{Experimental Setup}
    
  We employ three mainstream pretrained LLMs of progressively increasing model sizes to validate the scaling effects. These models include Llama-3-8B\footnote{\url{https://llama.meta.com/llama3/}}, Mistral-7B-Instruct-v0.1 \citep{jiang2023mistral}, and Gemma-2B \citep{team2024gemma}. For the hyperparameters of WEPO, we set the deviation parameter \(\beta\) to 0.95 and the negative sample ratio to 1:3. In Mind2Web \citep{deng2024mind2web}, a DeBERTa \citep{he2020deberta} model trained within the candidate generation module uses recall@50 for ranking elements and constructing a candidate pool for subsequent experiments. We similarly select a pruning ratio of \(k = 50\) to maintain consistency for comparison, preserving 50 central elements and their neighboring elements tagged with \texttt{element\_ID}. For the selection of \( k \) and \( n \) values, we provide detailed explanations in the forthcoming ablation studies. All models were configured with a maximum context length of 8192 tokens. We employ the Low Rank Adaptation (LoRA) technique \citep{hu2021lora} for parameter-efficient fine-tuning, which helps reduce memory usage and conserve budget. The learning rate is set at $0.0001$, and we use a combination of learning rate warmup and a cosine decay strategy for training. Additional training details can be found in \hyperref[sec:detail]{Appendix D}.
    
  \textbf{Evaluation Metrics.} In this paper, we adopt the step success rate (SSR) and Operation F1 score from \citet{deng2024mind2web}. We no longer use element accuracy and success rate because it is evident that both metrics are linearly associated with SSR, and successful interaction for the web navigation agent is only considered when both the element positioning and the corresponding operation are correct, which is precisely what SSR measures. The Operation F1 score is equally indispensable as it considers the accuracy of the input value for \texttt{Type} and \texttt{SELECT} commands. Furthermore, both baseline studies by \citet{gur2023real} and \citet{hong2023cogagent} exclusively utilized SSR as the sole metric.
    
  Additionally, we designed the element distance metric to measure the positional deviation between the elements selected by the WEPO model and the labeled elements. Consistent with the previous sampling strategy, we calculate the sum of the distances (in terms of steps) between the nodes corresponding to two different web elements and their lowest common ancestor (LCA) in the DOM tree to represent their relative positions.

  \begin{table*}
      \centering
      \begin{tabular}{p{0.38\textwidth}|p{0.11\textwidth}p{0.11\textwidth}p{0.11\textwidth}p{0.11\textwidth}}
        \toprule
        \textbf{Model}  SSR / Op. F1 (\%) & \textbf{overall} & \textbf{cross\_domain} & \textbf{cross\_task} & \textbf{cross\_website} \\
        \midrule
        Flan-T5-XL \textit{MindAct} & 43.5 / 69.1 & 39.6 / 66.5 & 52.0 / 75.7 & 38.9 / 65.2 \\
        Llama3-8B \textit{MindAct} & 55.1 / 65.8 & 57.6 / 68.7 & 56.1 / 63.2 & 51.7 / 65.6 \\
        \midrule
        CogVLM-17B \textit{CogAgent} & 58.2 / - & 59.4 / - & 62.3 / - & 54.0 / - \\
        HTML-T5-XL + Flan-U-PaLM \textit{WebAgent} & 49.7 / - & 48.3 / - & 57.8 / - & 42.9 / - \\
        \midrule
        Llama3-8B \textit{WEPO random} & 61.1 / 73.9 & 62.5 / 77.5 & 62.5 / 67.2 & 58.4 / \textbf{77.1} \\
        Mistral-7B \textit{WEPO random} & 57.2 / 73.8 & 58.0 / 75.9 & 59.0 / 72.1 & 54.8 / 73.3 \\
        Gemma-2B \textit{WEPO random} & 45.4 / 49.5 & 49.1 / 55.7 & 45.2 / 42.9 & 41.9 / 50.0 \\
        \midrule
        Llama3-8B \textit{WEPO distance-based} & \textbf{63.5} / 76.1 & \textbf{64.4} / \textbf{81.6} & \textbf{66.1} / 74.9 & \textbf{60.0} / 71.9 \\
        Mistral-7B \textit{WEPO distance-based} & 59.5 / \textbf{76.8} & 59.8 / 80.9 & 62.1 / \textbf{76.2} & 56.7 / 73.2 \\
        Gemma-2B \textit{WEPO distance-based} & 48.4 / 53.3 & 53.5 / 60.3 & 47.9 / 48.8 & 43.7 / 50.7 \\
        \bottomrule
      \end{tabular}
      \caption{\label{model-comparison}
        Overall performance of various models on different test sets, with evaluation metrics corresponding to SSR / Operation F1 (\%). The results were obtained under a negative sample ratio of $1:3$. Notably, our \texttt{Llama3-8B-WEPO} model achieved the highest scores in both overall SSR and Operation F1. Our top scores exceeded those of the much larger CogAgent (17B) model by $5.3\%$ and WebAgent (3B + 540B) by $13.8\%$. Additionally, our smaller \texttt{Gemma-2B-WEPO} model managed to closely match and even slightly outperform the approximately 3B Flan-T5 based models \citep{chung2024scaling} like MindAct.
      }
    \end{table*}
  
  \subsection{Results}

  We thoroughly evaluate WEPO on the partitioned three-tier held-out test sets in Mind2Web \citep{deng2024mind2web}, including \textbf{cross-domain}, \textbf{cross-website} and \textbf{cross-task} datasets. This allows us to understand how well our method can generalize across different domains, websites, and tasks.
  
  The experimental results are detailed in Table \ref{model-comparison}. Compared to previous works \citep{deng2024mind2web,hong2023cogagent,gur2023real}, without utilizing any multimodal information or customized model architecture, our best results obtained by fine-tuning the 8-billion parameter Llama-3 pretrained model with web element preference learning exceeded the three baselines by $20.0\%$, $5.3\%$, and $13.8\%$, respectively. Even though there is still a significant gap between the model performance of gemma-2B and the larger model ($>10\%$), these results demonstrate the effectiveness of the WEPO method, proving its efficiency and generalizability in enabling LLM-based agents for web navigation tasks. In assessing generalization capabilities at different levels, although \citet{deng2024mind2web} found that their model performed best in the Cross-Task setting, WEPO has narrowed the gap between different test sets to less than 6.1\%. Additionally, observing the performance of WEPO models of different sizes, we found that the step success rate increases with model size, which verifies the presence of the scaling effect in the Mind2Web real-world benchmark.
  
  To eliminate the influence of model base choice on the results, we thoroughly applied supervised fine-tuning using the action prediction prompts from Mind2Web on Llama3-8B. The results indicate that compared to this upgraded MindAct baseline, WEPO also outperforms by 8.4\% on the SSR metric and by 11.0\% on the Operation F1 score. This again demonstrates that fine-tuning with positive human annotation only, which is a simple maximum likelihood approach, is less effective than our WEPO method. We then conducted a comparison of its performance with random sampling in Table \ref{model-comparison} to validate the effectiveness of the distance-based sampling method. Compared to random sampling, the distance-based approach achieved a 2.3\% to 3\% higher SSR, demonstrating its efficiency. In the subsequent analysis of the element distance evaluation metric, it can also be seen that this method successfully enhances the accuracy of the model's selections, indirectly suggesting better alignment with user high-level intents.
  
  \textbf{What exactly is the role of preference learning in enhancing performance?} For the Operation F1 score, we observed a significant improvement from Gemma-2B-WEPO to Mistral-7B-WEPO, which indicates that an increase in the number of LLM training parameters not only makes the web element localization more accurate but primarily enhances the accuracy of input or select text values. For MindAct, however, our best performance was not significantly ahead in F1 scores, suggesting that the WEPO method, compared to traditional supervised fine-tuning, enhances the accuracy of element selection in pretrained LMs. We infer that WEPO leverages this enhancement through a contrastive training scheme, wherein the model learns to distinguish between elements that are critical for decision-making and those that are not. Particularly when intentions are abstract, a web page at any given moment may contain multiple elements that are easily confused, and WEPO reduces the likelihood of incorrect element selection by the agent. By incorporating a contrastive mechanism, WEPO not only improves the accuracy of navigation tasks but also enhances the model’s generalizability across different web layouts and designs. 
  
\begin{figure}[t]
    \centering
    \begin{minipage}[t]{0.48\linewidth}
        \includegraphics[width=\linewidth]{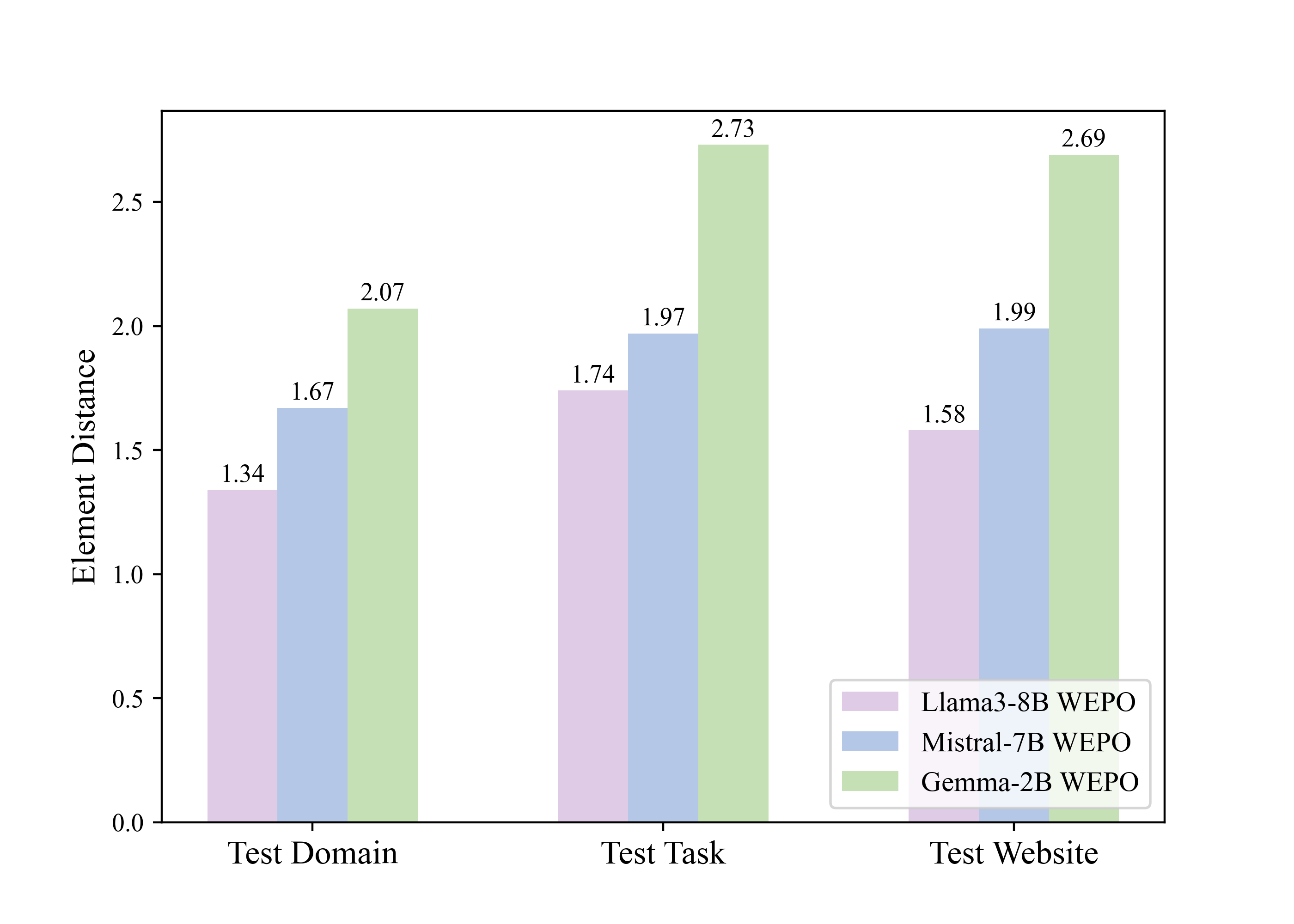}
        \caption{Statistical distribution of Element Distance for different models (Llama3-8B, Mistral-7B and Gemma-2B) on the test dataset. As the model size increases, the relative deviation in element distances decreases.}
        \label{fig:id_distance}
    \end{minipage}
    \hfill
    \begin{minipage}[t]{0.48\linewidth}
        \includegraphics[width=\linewidth]{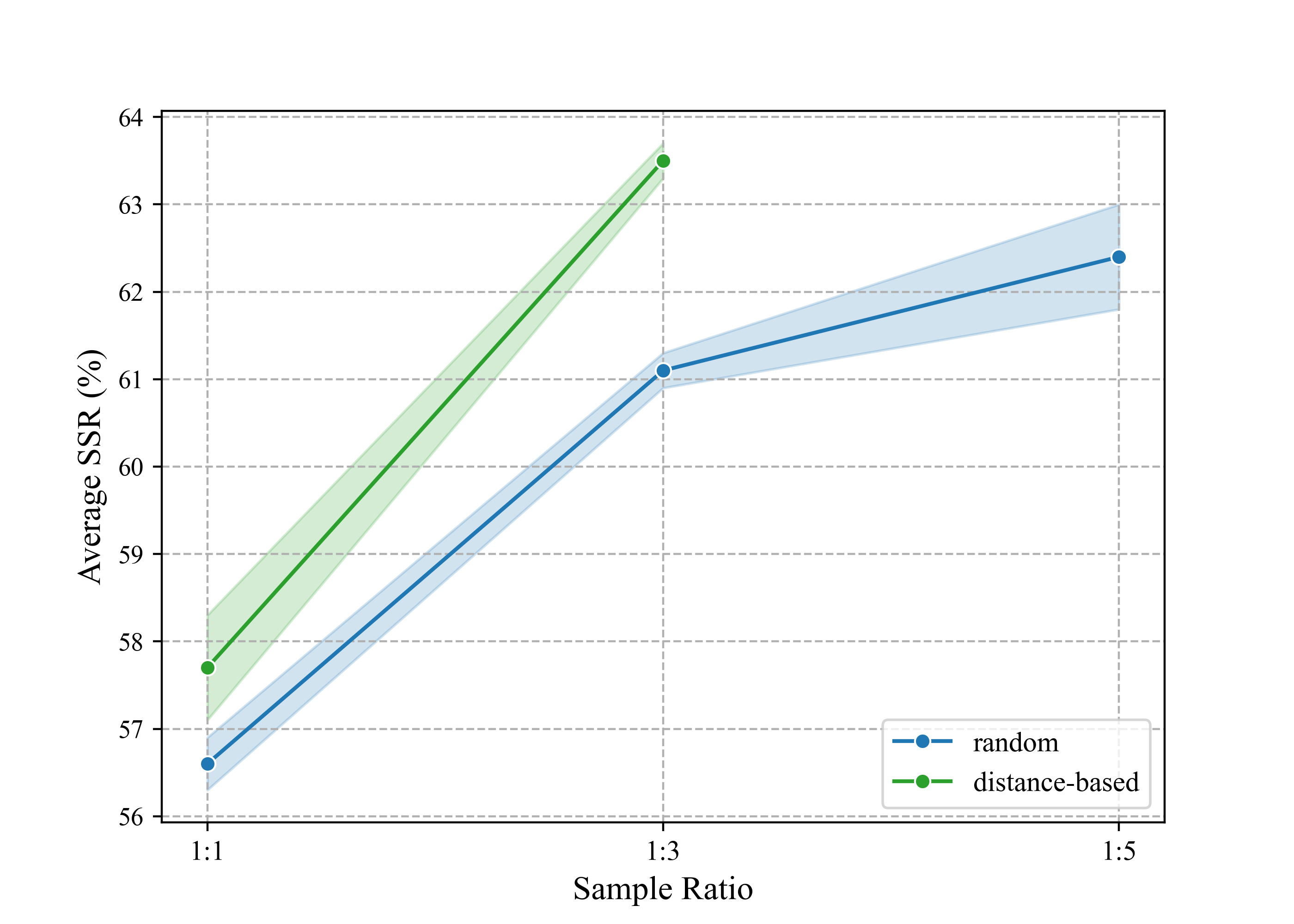}
        \caption{Ablation studies on the negative ratio. We experimented with the Llama3-8B-WEPO model at $n = 1, 3, 5$ and calculated the average SSR (\%) on three cross-test sets, which were $57.7\%$, $63.5\%$ for distance-based sampling and $56.6\%$, $61.1\%$, and $62.4\%$ for random sampling, respectively. An elbow point was observed at $n = 3$ for random sampling, where the increase in SSR sharply levels off. Furthermore, the performance of distance-based sampling at a 1:3 ratio has already surpassed that of random sampling at a 1:5 ratio by 1.1\%.}
        \label{fig:neg_ratio}
    \end{minipage}
\end{figure}

  The results in Figure \ref{fig:id_distance} provide great support for this inference. As the model size increases, the element distance consistently decreases. Element distance, which represents the relative position of elements in the DOM, is closely correlated with the elements' function and design purpose. Coupled with the score analysis in Table \ref{model-comparison}, this decreasing deviation suggests that the Llama3-8B-WEPO model becomes increasingly accurate in aligning with the intended functions and design of web elements compared to the smaller models. This result clearly indicates that the model has successfully learned to recognize these web design differences, proving that WEPO is an effective method for adapting to the HTML structure, or more broadly for aligning with web design principles. Furthermore, we have demonstrated the effectiveness of our newly proposed element distance evaluation metric.

  \textbf{Why choose a 1:3 ratio for negative samples?} We also conducted ablation experiments on the negative sample ratio in the WEPO Algorithm \ref{alg:Framework}. Empirically, we aimed to sample as many negative samples as possible to enhance performance through WEPO without over-sampling and excessively increasing the training overhead. We uniformly sampled $n$ values of $1$, $3$, and $5$, selected the best-performing Llama3-8B-WEPO for experimentation and averaged the results from two rounds. As shown in Figure \ref{fig:neg_ratio}, the overall average scores on the test set increase with larger $n$ values, with a noticeable elbow point at $n=3$ for random sampling. We ultimately selected a 1:3 ratio as a fixed hyperparameter to avoid linearly increasing training costs. Additionally, we observed that the distance-based sampling method at $n=3$ outperformed random sampling at $n=5$, significantly improving overall sample learning efficiency during training. Furthermore, too many negative samples could potentially create an imbalance between positive and negative sample quantities. However, we did not experiment with larger $n$ values to verify potential performance degradation, as excessively large $n$ values could hinder reproducibility. 
  
  \textbf{What is the impact of HTML $k$-pruning on performance?} To address this question, we conducted ablation studies on the selection of $k$, with results shown in Table \ref{tab:k_ablation}. As our pruning is based on the first stage ranking LM of MindAct, understanding the impact of this preprocessing module is crucial. The authors of \citet{deng2024mind2web} disclosed that when $k=50$, the fine-tuned DeBERTa \citep{he2020deberta} model achieved recall accuracies of $88.9\%$, $85.3\%$, and $85.7\%$ on three held-out test datasets, and smaller values of $k$ were not adopted due to lower recall rates. 

  \begin{table*}
    \centering
    \begin{tabular}{l|c|c|c}
      \toprule
      \textbf{Model} SSR (\%) & \textbf{cross\_domain} & \textbf{cross\_task} & \textbf{cross\_website} \\
      \midrule
      Flan-T5-XL MindAct $k=50$ & 39.6 & 52.0 & 38.9 \\
      Llama3-8B MindAct $k=50$ & 57.6 & 56.1 & 51.7 \\
      Llama3-8B \textit{WEPO} $k=50$ & 64.4 & 66.1 & 60.0 \\
      Llama3-8B \textit{WEPO} $k=10$ & 49.7 & 55.0 & 46.5 \\
      \midrule
      Llama3-8B \textit{WEPO} $k=10$ w. ground truth $\hat{e}$ & 87.2 & 88.7 & 85.4 \\
      \bottomrule
    \end{tabular}
    \caption{\label{tab:k_ablation}
      SSR performance (\%) from ablations on the cleaned HTML pruning ratio $k$ value. For WEPO training, we forcibly included the ground truth element, akin to a teacher-forcing mechanism. The results shown in the table, except for the last row, represent the SSR after reassembling HTML from elements filtered through the first-stage ranking. Llama3-8B-WEPO maintained a $10.1\%$ improvement to Flan-T5 based MindAct even after reducing $k$.
    }
  \end{table*}
  
  However, in our results, when we set $k$ to a smaller value of $10$, Llama3-8B-WEPO still performed above the baseline. We deduce that although reducing $k$ significantly decreases the recall rate of the ranking LM, it provides the WEPO-trained model with a shorter HTML snippet and fewer ID options, enhancing discrimination accuracy. Moreover, when we forcibly added the ground truth web element $\hat{e}$ directly into the candidate pool, the ablation model's SSR surged to over 80\%. This conclusively demonstrates that the WEPO method has significantly improved the model's capability in action prediction and has shifted the original candidate retrieval module from a secondary issue to a major bottleneck limiting the web agent's progress on complex long-context web pages.
  
  \section{Future Work}
  There remain several avenues for further development of this approach. While WEPO has shown empirical success, a deeper theoretical analysis could provide more foundational insights into why and how preference optimization effectively enhances web navigation tasks. Inspired by the advancements of models like WebFormer \citep{wang2022webformer} and HTML-T5 \citep{gur2023understanding}, future iterations of WEPO could benefit from a dedicated HTML encoder that is specifically tailored to understand the hierarchical and nested structures of web documents. 
  
  The current implementation of WEPO has not explored its potential over extremely large context lengths such as \texttt{phi-3-128k} \citep{abdin2024phi}. Future research could look into the scaling abilities of WEPO when applied to such models, which might be crucial for handling complex web navigation tasks that involve detailed web pages.
  
  While WEPO shows promising results in a controlled benchmark environment, its ability to generalize across highly diverse, real-world web interfaces remains an area for further investigation. The variability in web design, interactive elements, and underlying technologies across different websites may affect the consistency of WEPO's performance. We plan to test the performance of the WEPO method on additional mainstream benchmarks such as WebLINX and WebVoyager \citep{lu2024weblinx,he2024webvoyager} in future work.
  
  \section{Conclusion}
  
  This paper introduced the Web Element Preference Optimization (WEPO), a simple yet novel framework that integrates Direct Preference Optimization (DPO) into LLM-based web navigation tasks. WEPO enhances LLM performance by leveraging distance-based non-salient HTML elements for contrastive learning, effectively aligning model operations with user intent. Our empirical evaluations on the Mind2Web benchmark demonstrate that WEPO surpasses traditional models, achieving an 13.8\% improvement over the WebAgent baseline and a 5.3\% enhancement beyond the visual language model CogAgent, while also exhibiting strong generalization across diverse web environments. This research significantly enhances the capabilities of LLMs in web navigation task, contributing to more efficient and intuitive web interactions.

\newpage

\bibliography{wepo}

\newpage

\section*{Appendix} 

\subsection*{A Broader Impact}
\label{sec:appendix_a}

In implementing the Web Element Preference Optimization (WEPO) model for autonomous web navigation, we prioritized ethical considerations, maintaining privacy by anonymizing web data. All tests were conducted under human supervision to monitor and address any undesired behaviors, setting a standard for responsible AI deployment in web environments.

\subsection*{B WEPO Hypothesis}
\label{sec:hypo}

The efficacy of web navigation tasks heavily relies on the ability to accurately interpret and act upon user intent. However, previous methods often overlook the latent potential of HTML elements that are not directly associated with the user’s immediate goals, thereby under-utilizing the rich semantic structure of web pages. Leveraging current preference learning methods, we aim to utilize non-salient elements to construct negative samples for introducing contrastive learning.

However, the viability of this approach is predicated on a hypothesis that at a given moment \( t \), within \( s_t \) corresponding to \( E_t \), there is exactly one ground truth element \( \hat{e_t} \) that can ultimately achieve the user intent. To illustrate the limitation of this hypothesis, consider an example. If you want to search for images of apples on Google, you might first type \textit{apple} in the search bar and press enter, then click the image button; alternatively, you could click the image button in the upper right corner first, then search for \textit{apple}. These two trajectories are equivalent, and we refer to this situation as \textbf{Equivalent Multiple Paths (EMPs)}. Thus, the WEPO hypothesis assumes the non-existence of EMPs in the web environment, which simplifies the problem setting and may slightly reduce the selection diversity.

While WEPO may not learn with as much diversity across multiple equivalent paths, neither do the baselines based on vanilla supervised methodology. The best way to address this issue is by improving dataset quality through improving data annotation strategies. Additionally, we conducted straightforward human evaluations on the diversity of different sampling results, indicating that the elements sampled from a page (random or distance-based) are almost entirely independent of the target elements (\(p<0.01\)). Thus, empirically, WEPO does not compromise diversity that much and significantly enhances task performance.

\begin{figure}[t]
    \centering
    \includegraphics[width=0.5\linewidth]{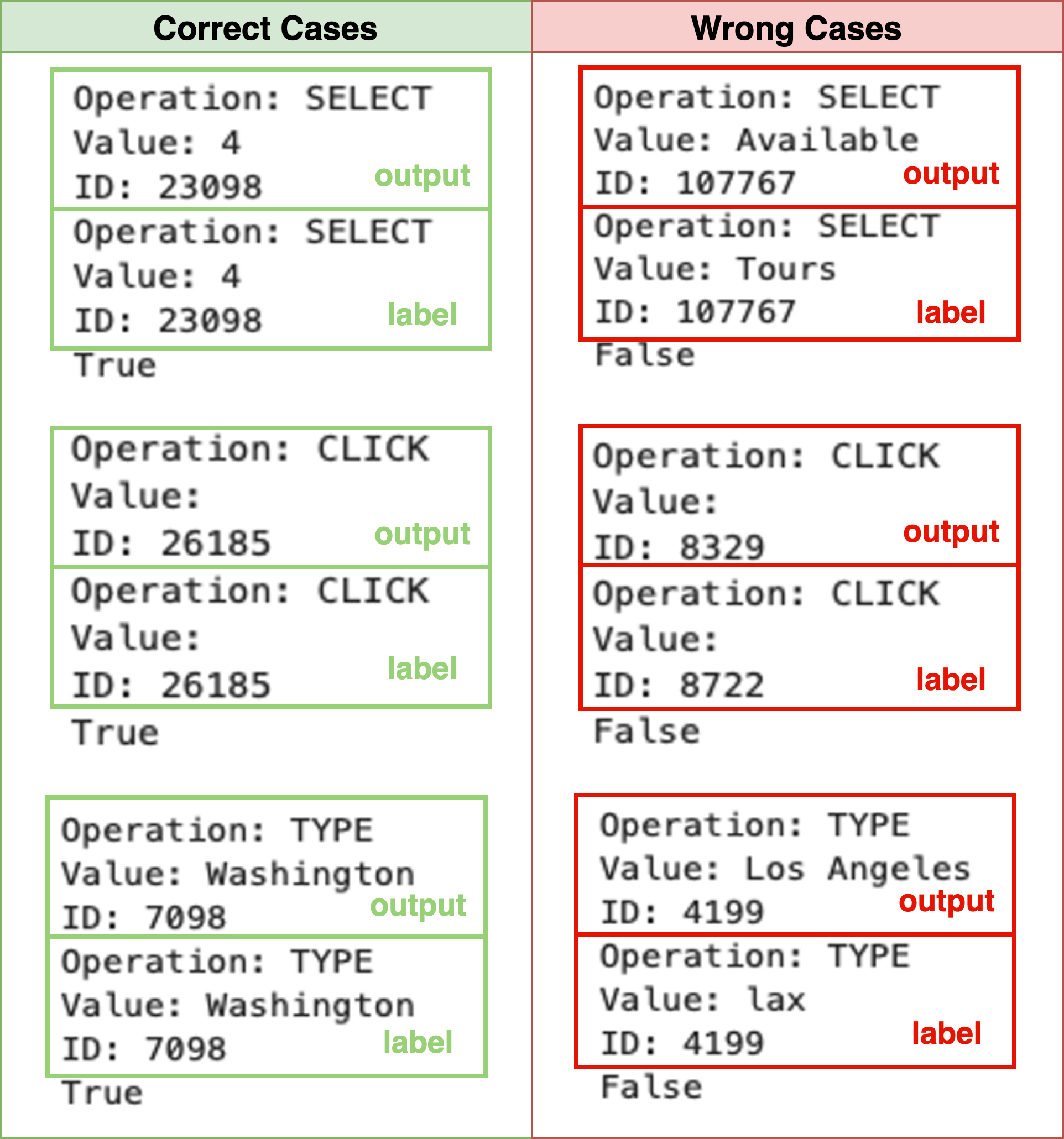}
    \caption{A collection of correct and incorrect actions generated by WEPO models. The correct cases are within the green boxes on the left, and the incorrect cases are within the red boxes on the right. It can be seen that for correct generations, if the action is a \texttt{CLICK}, the element ID must be the same, and for \texttt{TYPE} and \texttt{SELECT}, the textual value must also be identical. Therefore, the common causes of errors include incorrect element locating or inconsistencies in textual content.}
    \label{fig:cases}
\end{figure}
 
\subsection*{C Case Study}
\label{sec:case}

To more clearly demonstrate the action schemas by the LLM and the possible failure scenarios, this section presents six selected generative results, as shown in Figure \ref{fig:cases}. It is evident that for the LLM fine-tuned with WEPO, the correct outputs often achieve accurate element localization and completely consistent textual input, demonstrating its ability of intent-based reasoning in long HTML contexts. The possible errors arise from inaccurately selecting other web elements or failing to generate the corresponding textual value.

\subsection*{D Training Details}
\label{sec:detail}

The training process is facilitated by the open-source training frameworks Llama-Factory\footnote{\url{https://github.com/hiyouga/LLaMA-Factory}} and Unsloth\footnote{\url{https://github.com/unslothai/unsloth}} for DPO training, which offer convenient deployment options. We conducted model training on four 48GB L20 GPUs, and Table \ref{tab:model_performance} provides an overview of the approximate training and testing durations. During the training process, we set the validation set ratio to $15\%$ and the number of epoch to $1$. We utilized the Low Rank Adaptation (LoRA) technique, setting the rank $r=64$. We set the deviation parameter to $\beta=0.95$. The learning rate was set to $0.0001$, and the warm-up step ratio was set to $0.1$.

\begin{table*}[h]
  \centering
  \begin{tabular}{p{0.3\linewidth}|p{0.17\linewidth}|p{0.17\linewidth}}
    \toprule
    \textbf{Models} & \textbf{Training} & \textbf{Inference} \\
    \midrule
    Meta-Llama-3-8B-Instruct & 7 hours & 2 hours \\
    Mistral-7B-Instruct-v0.1 & 7 hours & 2 hours \\
    gemma-2b-it & 3 hours & 1 hour \\
    \bottomrule
  \end{tabular}
  \caption{Comparison of different pretrained models training and inference times.}
  \label{tab:model_performance}
\end{table*}

\newpage
\subsection*{E Statistics on Mind2Web}
\label{sec:stat}

\begin{figure*}[t]
  \includegraphics[width=\linewidth]{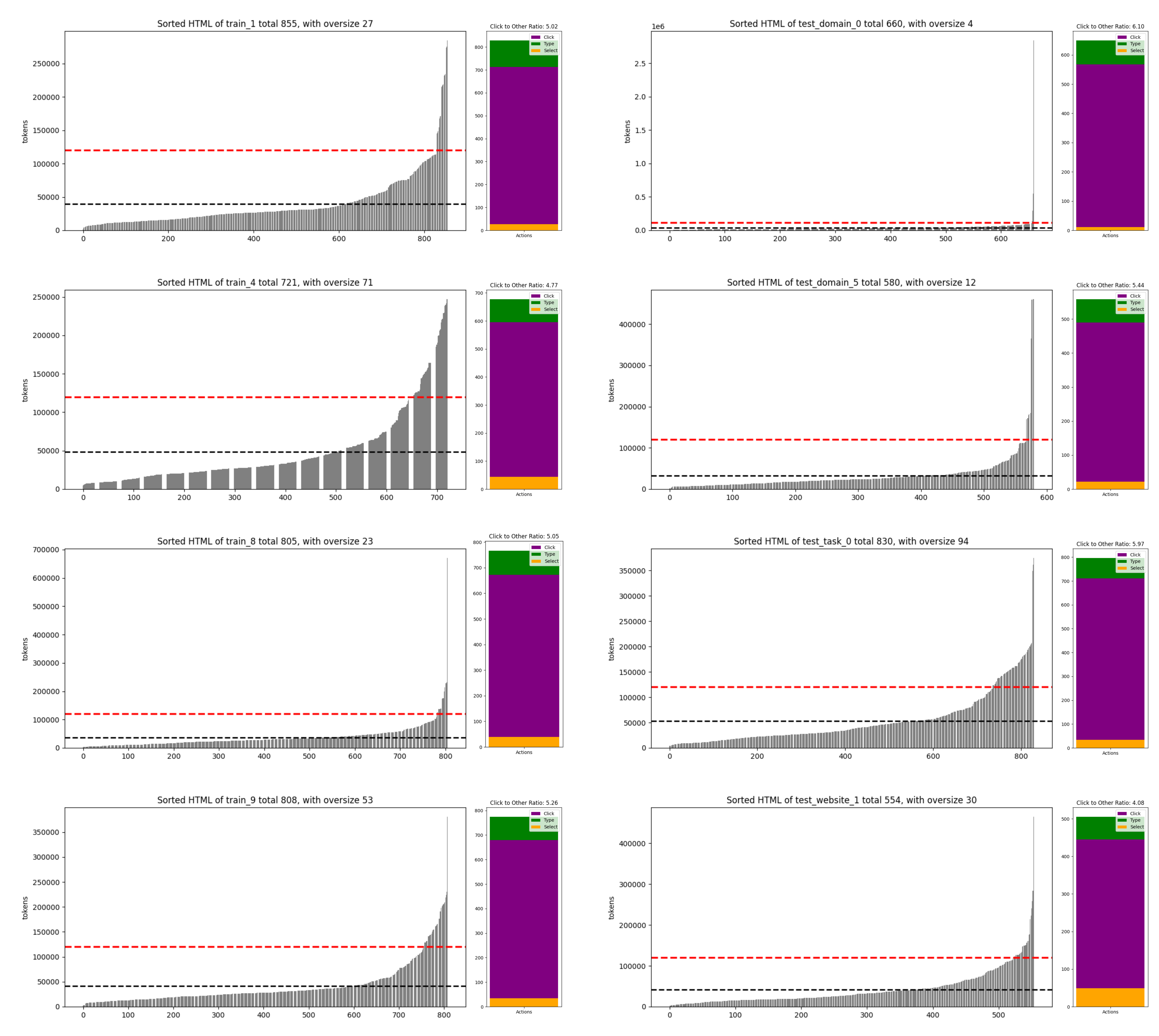}
  \caption{Statistical visualization of HTML snippet lengths and predefined action proportions in the Mind2Web dataset. We randomly selected eight subsets for display. The red dashed line represents the upper limit of the context window length for mainstream open-source LLMs at 128k tokens, while the gray dashed line indicates the average token length of HTML snippets. In the proportional bar charts on the right, purple corresponds to \texttt{CLICK}, green to \texttt{TYPE}, and yellow to \texttt{SELECT}.}
  \label{fig:mind2web}
\end{figure*}

\begin{figure*}[t]
  \includegraphics[width=\linewidth]{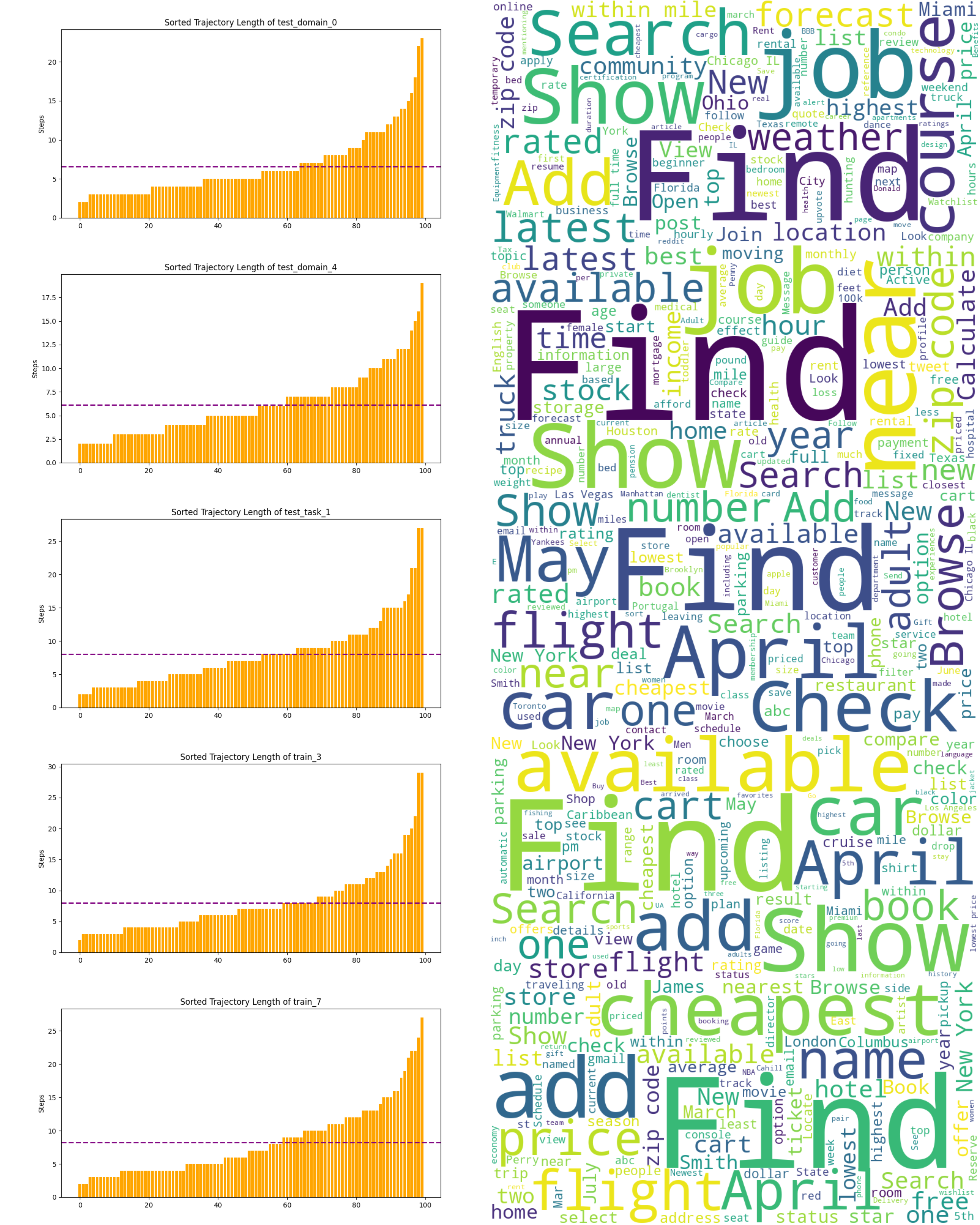}
  \caption{Statistical visualization of data trajectory lengths and high-frequency words of intents in the Mind2Web dataset. We randomly selected five subsets for display. The purple dashed line represents the average length of the trajectories, and in the word cloud, the larger the word, the higher its frequency of occurrence.}
  \label{fig:cloud}
\end{figure*}

To better analyze the experimental results, we visualized the sample distribution of the Mind2Web dataset. As shown in Figure \ref{fig:mind2web}, we statistically analyzed the average token length of HTML across the entire dataset and the proportion of different actions, with some samples randomly selected for display. The sections exceeding the red dashed line represent HTML snippets with token lengths surpassing 128k, which are challenging for most currently available large language models to process. The average length of an HTML snippet is indicated by a gray line. Therefore, preprocessing and pruning are inevitable for current web navigation LLMs. In the proportional bar charts on the right side of each subplot, the purple block corresponding to the \texttt{CLICK} action constitutes the highest proportion, while the remaining two actions requiring optional textual values have lower proportions. The ratio of \texttt{CLICK} actions to other actions ranges from $4$ to $6$, indicating that using only the Step Success Rate as a metric is insufficient. The Operation F1, calculated specifically for \texttt{TYPE} and \texttt{SELECT} actions, serves as a robust complement.

Figure \ref{fig:cloud} primarily visualizes the ground truth operation trajectory lengths and high-frequency words in intent descriptions within the Mind2Web dataset. On the left side of the figure, it can be seen that the longest interaction trajectories can exceed 20 steps, while the average number of steps ranges between 5 and 10. This indicates that web navigation tasks require LLM agents to possess robust long-sequence reasoning and memory capabilities to correctly select logically coherent actions based on past clicks or input actions. The right side presents a word cloud based on all intents, revealing that verbs such as \textit{find}, \textit{show}, and \textit{search} are most common, along with frequently occurring nouns like \textit{price}, \textit{car}, \textit{course}, and adjectives like \textit{latest}, \textit{cheapest}, which are typical components of concise natural language web intents.

\subsection*{F WEPO Example Flow}
\label{sec:flow}

\begin{figure*}[t]
  \includegraphics[width=\linewidth]{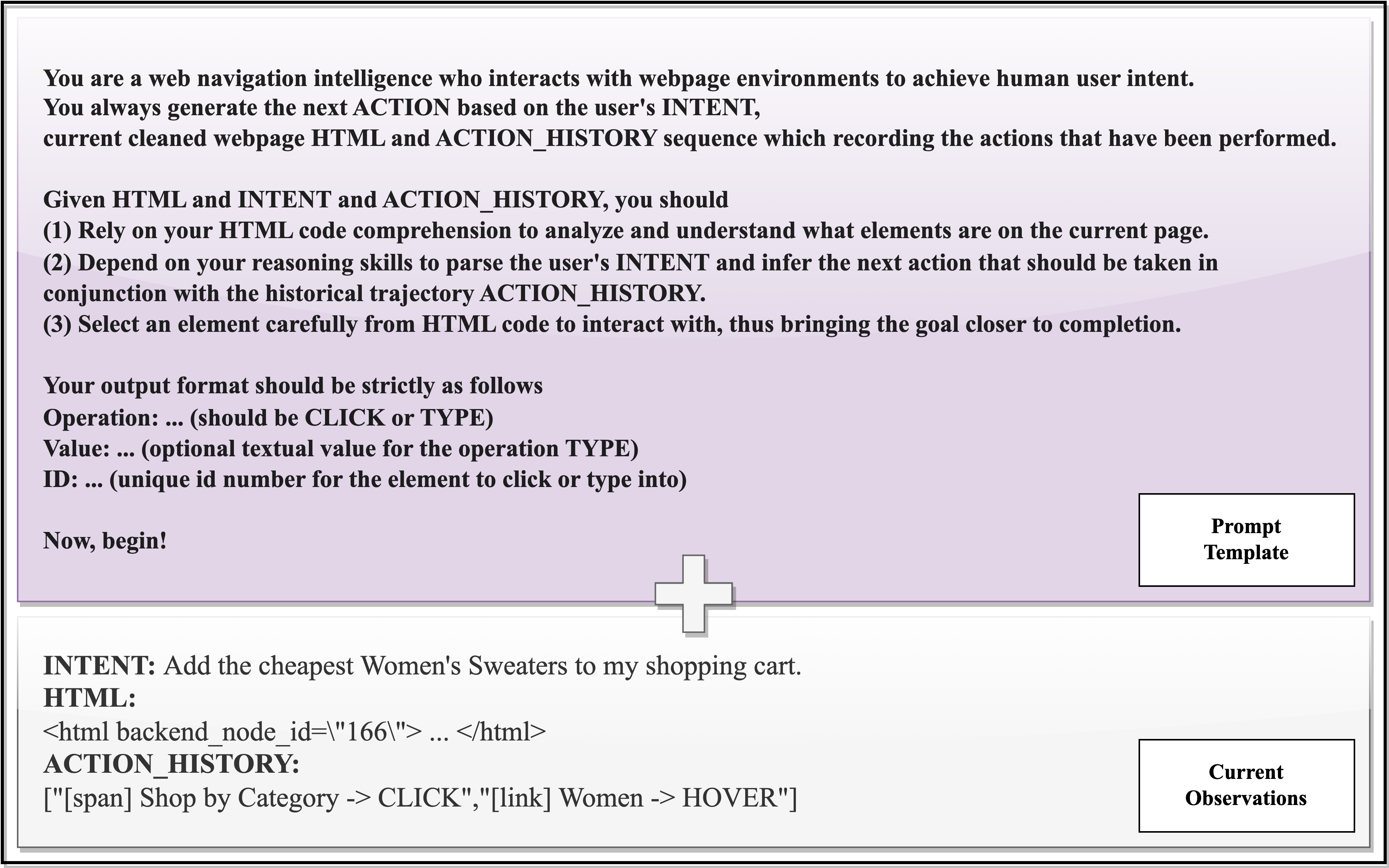}
  \caption{An example flow of WEPO's single-step reasoning. In the upper part of the prompt, the prompt template primarily facilitates identity setting, instruction explanation, and output format specification. In the lower part of the prompt, the current conversation directly puts all the information together, which comprises three primary sections: the user task objective, the sequential history of actions, and the HTML snippet of the current web page.
}
  \label{fig:flow}
\end{figure*}

In this section, we present the design of our prompt template. During the design process, we drew upon current mainstream prompt writing techniques, including identity setting, step-by-step instruction breakdown, and output format standardization. See Figure \ref{fig:flow} for details.

\bibliographystyle{unsrtnat}






\end{document}